\pdfoutput=1

\documentclass[11pt]{article}

\usepackage[table]{xcolor}
\usepackage{pgfplotstable}

\pgfplotstableset{
    color cells/.style={
        col sep=comma,
        string type,
        postproc cell content/.code={%
                \pgfkeysalso{@cell content=\rule{0cm}{2.4ex}\cellcolor{red!##1}\pgfmathtruncatemacro\number{##1}\ifnum\number>50\color{white}\fi##1}%
                },
        columns/x/.style={
            column name={},
            postproc cell content/.code={}
        }
    }
}
\usepackage{emnlp2021}

\usepackage{times}
\usepackage{latexsym}

\usepackage[T1]{fontenc}

\usepackage[utf8]{inputenc}

\usepackage{microtype}

\usepackage{graphicx}
\usepackage{array}
\usepackage{url}

\usepackage{multirow}

\title{On the ability of monolingual models to learn language-agnostic representations}

\author{Leandro Rodrigues de Souza,$^{1}$ 
\textbf{Rodrigo Nogueira,$^{1,2,3}$ Roberto Lotufo$^{1,2}$} \\
\\
$^{1}$ Faculty of Electrical and Computer Engineering, University of Campinas (UNICAMP)\\
$^{2}$ NeuralMind Inteligência Artificial\\
$^{3}$ David R. Cheriton School of Computer Science, University of Waterloo\\
}

\date{}

\begin{document}

\maketitle

\begin{abstract}
Pretrained multilingual models have become a \textit{de facto} default approach for zero-shot cross-lingual transfer. Previous work has shown that these models are able to achieve cross-lingual representations when pretrained on two or more languages with shared parameters. In this work, we provide evidence that a model can achieve language-agnostic representations even when pretrained on a \textit{single language}. That is, we find that monolingual models pretrained and finetuned on different languages achieve competitive performance compared to the ones that use the same target language. Surprisingly, the models show a similar performance on a same task regardless of the pretraining language. For example, models pretrained on distant languages such as German and Portuguese perform similarly on English tasks. 

\end{abstract}

\section{Introduction}

Pretrained language models, such as BERT \citep{bert} and T5 \citep{t5}, still rely on high quality labeled datasets for achieving good results in NLP tasks. Since creating such datasets can be challenging \citep{andbiswas2009complex, sabou2012}, many resource-lean languages suffer from the lack of data for training.

Multilingual pretraining has been developed to overcome this issue, giving rise to multilingual models like mBERT~\citep{bert}, XLM~\citep{Conneau2018, Conneau2020} and mT5~\citep{xue2021mt5}. Differently from monolingual models, they are pretrained with mixed batches from a wide range of languages. The resulting model finetuned on a task in a high-resource language often performs well on the same task in a different language never seen during finetuning. Such models thus exhibit the so called \textit{zero-shot cross-lingual transfer} ability.

This approach has become the de facto default language transfer paradigm, with multiple studies and benchmarks reporting high transfer performance \citep{Pires2019, xtreme_paper}. At first, it has been speculated that a shared vocabulary with common terms between languages would result in a shared input representation space, making it possible for the model to attain similar representations across languages \citep{Pires2019}. However, more recent work has shown that having a shared vocabulary is not critical for cross-lingual transfer \citep{Artetxe2019, Wu2020}.

\citet{Wu2020} have released a large study on bilingual models to identify the most important characteristics that enable cross-lingual transfer. The authors found that a shared vocabulary with anchor points (common terms in both languages) plays a minor role. Sharing the Transformers' parameters when training on datasets from two languages seems to be critical. They also found an interesting property in monolingual models: they have similar representations in the first layers of the network that can be aligned with a simple orthogonal mapping function. 

Similar findings have been presented by \citet{Artetxe2019}. They developed a procedure to learn representations for a specific language (i.e., \textit{word embedding layer}) aligned with pretrained English layers of a Transformer model. This enables the model to transfer knowledge between languages by only swapping its lexicon.

As it becomes clearer that models trained on different languages achieve similar representations and cross-lingual transfer can be reached with quite simple methods, we raise the following question: \textbf{Would monolingual models be able to learn from labeled datasets in a foreign language (i.e., a different language from its pretraining) without any adjustments at all?}

We ground our hypothesis on the fact that masked language modeling has demonstrated to achieve strong representations even in different domains \citep{lu2021pretrained, proglanguageagnostic}. To the best of our knowledge, our work is the first to experiment on monolingual models learning from a dataset in a different language.

Based on these observations, we design an experiment to test if monolingual models can leverage concepts acquired during pretraining and learn a new language and a new task from finetuning only. We pretrain a model on a source language and finetune it on a different language. We also evaluate models without pretraining, as a control experiment, to discard the hypothesis that the model is learning patterns only from the finetuning dataset.

\medskip
\noindent\textbf{Main contribution.} We demonstrate that monolingual models can learn a task in a foreign language. Our results show that, despite those models have inferior performance when compared to the standard approach (i.e., pretraining and finetuning a model on the same language), they can still learn representations for a downstream task and perform well when compared to models without pretraining. This raises the hypothesis that MLM pretraining provides the model some language-agnostic properties that can be leveraged regardless of the language of the task. 

In contrast to the current literature, the monolingual models used in this work do not rely on shared parameters during pretraining with multi-lingual datasets. They are pretrained, instead, on a single language corpus and finetuned on a task in a different (foreign) language. The results raise questions about language-agnostic representations during MLM pretraining and contribute to future directions in cross-lingual knowledge transfer research.


\section{Related Work}

Multilingual pretraining has been widely adopted as a cross-lingual knowledge transfer paradigm \citep{bert, Lample2019, Conneau2020, xue2021mt5}. Such models are pretrained on a multilingual corpus using masked language modeling objectives. The resulting models can be finetuned on high-resource languages and evaluated on a wide set of languages without additional training.

\citet{Pires2019} first speculated that the generalization ability across languages is due to the mapping of common word pieces to a shared space, such as numbers and web addresses.

This hypothesis was tested by \citet{Artetxe2019}. The experiment consisted of pretraining an English BERT model's lexicon in a target language dataset: only \textit{word embeddings} are trainable. The authors, then, have transferred knowledge by taking a BERT model finetuned on English and swapping its lexicon to the desired language. They obtained results similar to those obtained by \citet{Pires2019}. Therefore, they confirmed the hypothesis that the model can learn representations that are language-agnostic and that do not depend on a shared vocabulary nor joint pretraining.

\citet{Wu2020} have also demonstrated that a shared vocabulary with anchor points contributes little to language transfer. In their experiments, the authors carried out an extensive analysis on the results of bilingual models to identify the factors that contribute most to language transfer. The results indicate that the main component is related to the sharing of model parameters, achieving good results even in the absence of parallel \textit{corpus} or shared vocabulary. They also found that monolingual models pretrained on MLM have similar representations, that could be aligned with a simple orthogonal mapping.

\citet{zhao2020inducing} observed that having a cross-lingual training objective contributes to the model learning cross-lingual representations, while monolingual objectives resulted in language-specific subspaces. These results indicate that there is a negative correlation between the universality of a model and its ability to retain language-specific information, regardless of the architecture. Our experiments show that a single language pretraining still enables the model to achieve competitive performance in other languages via finetuning (on supervised data) only.

\citet{libovicky2019language} found that mBERT word embeddings are more similar for languages of the same family, resulting in specific language spaces that cannot be directly used for zero-shot cross-lingual tasks. They also have shown that good cross-lingual representations can be achieved with a small parallel dataset. We show, however, that a monolingual model can achieve competitive performance in a different language, without parallel corpora, suggesting that it contains language-agnostic properties.

While multilingual models exhibit great performance across languages \citep{xtreme_paper}, monolingual models still perform better in their main language. For instance, \citet{monolingual-crosslingual-ms2019} distill a monolingual model into a multilingual one and achieve good cross-lingual performance. We take on a different approach by using the monolingual model itself instead of extracting knowledge from it.

\citet{rust2020good} compared multilingual and monolingual models on monolingual tasks (i.e., the tasks whose language is the same as the monolingual model). They found that both the size of pretraining data in the target language and vocabulary have a positive correlation with monolingual models' performance. Based on our results, we hypothesize that a model pretrained with MLM using a large monolingual corpus develops both language-specific and language-agnostic properties, being the latter predominant over the former.

\section{Methodology}

Our method consists of a pretrain-finetune approach that uses different languages for both. We call \textbf{source language} as the language used for pretraining our models. We refer to \textbf{target language} as a second language, different from the one used for pretraining our model. We apply the following steps:

\begin{enumerate}
    \item Pretrain a monolingual model on the \textbf{source language} with masked language modeling (MLM) objective using a large, unlabeled dataset.

    \item Finetune and evaluate the model on a downstream task with a labeled dataset in the \textbf{target language}.

\end{enumerate}

The novelty of our approach is to perform a cross-lingual evaluation using \textit{monolingual models} instead of bi-lingual or multi-lingual ones. We aim to assess if the model is able to rely on its masked language pretraining to achieve good representations for a task even when finetuned on a different language. If successful, this would suggest that MLM pretraining provides the model with representations for more abstract concepts rather than learning a specific language.

\medskip
\noindent\textbf{Pretraining data}. Our monolingual models are pretrained on a large unlabeled corpus, using a source language's vocabulary. Some high-resource languages, such as English, have a high presence in many datasets from other languages, often created from crawling web resources. This may influence the model's transfer ability because it has seen some examples from the foreign language during pretraining. However, the corpora used to pretrain our models have a very small amount of sentences in other languages. For instance, Portuguese pretraining corpus has only 14,928 sentences (0.01\%) in Vietnamese.

\medskip
\noindent\textbf{Control experiment}. To discard the hypothesis that the monolingual model can learn patterns from the finetuning dataset, instead of relying on more general concepts from both finetuning and pretraining, we perform a control experiment. We train the models on the target language tasks without any pretraining. If models with monolingual pretraining have significantly better results, we may conclude that it uses knowledge from its pretraining instead of only learning patterns from finetuning data.

\medskip
\noindent\textbf{Evaluation tasks}. We follow a similar selection as in \citet{Artetxe2019} and use two downstream types of tasks for evaluation: natural language inference (NLI) and question answering (QA). Even though a classification task highlights the model's ability to understand the relationship between sentences, it has been shown that the model may learn some superficial cues to perform well \citep{gururangan-etal-2018-annotation}. Because of that, we also select question answering, which requires natural language understanding as well.

\section{Experiments}

In this section, we outline the models, datasets and tasks we use in our experiments.

\subsection{Models}

\begin{table*}
\centering
    \begin{tabular}{l c c r r r}
    \hline
    \textbf{Model name} & \textbf{Initialization} & \textbf{Language} & \textbf{\# of params.} & \textbf{Data size} \\
    \hline
    BERT-EN \citep{bert} & Random & English (en) & 108 M & 16 GB \\
    BERT-PT (\textit{ours}) & Random & Portuguese (pt) & 108 M & 16 GB \\
    BERT-DE \citep{german-bert} & Random & German (de) & 109 M & 12 GB\\
    BERT-VI \citep{phobert} & Random & Vietnamese (vi) & 135 M & 20 GB \\
    \hline
    \end{tabular}
    \caption{Pretrained models used in this work.}
    \label{table:models}
\end{table*}

We perform experiments with four base models, highlighted in Table~\ref{table:models}. The experiments run on the Base versions of the BERT model \citep{bert}, with 12 layers, 768 hidden dimensions and 12 attention heads. We use models initialized with random weights. We also report the number of parameters (in millions) and the pretraining dataset sizes in Table~\ref{table:models}.

\subsection{Pretraining procedure}

The selected models are pretrained from random weights using a monolingual tokenizer, created from data in their native language. We also select models that have been pretrained using the Masked Language Modeling (MLM) objective as described by \citet{bert}.

\subsection{Finetuning procedure}

We use the AdamW optimizer~\citep{adamwpaper} and finetune for 3 epochs, saving a checkpoint after every epoch. Results are reported using the best checkpoint based on validation metrics. The initial learning rate is \(3e-5\), followed by a linear decay with no warm-up steps.

For NLI, we use a maximum input length of 128 tokens and a batch size of 32. For QA, except for BERT-VI, our input sequence is restricted to 384 tokens with a document stride of 128 tokens and a batch size of 16. For BERT-VI, we use an input sequence of 256 tokens and a document stride of 85. This is due to the fact that BERT-VI was pretrained on sequences of 256 tokens, resulting in positional token embeddings limited to this size.

We have chosen this hyperparameter configuration based on preliminary experiments with English and Portuguese models trained on question answering and natural language inference tasks in their native languages. We do not perform any additional hyperparameter search.

For the control experiments, we finetune the models for 30 epochs using the same hyperparameters, with a patience of 5 epochs over the main validation metric. This longer training aims to discard the hypothesis that the pretrained models have an initial weight distribution that is favorable compared to a random initialization, which could be the main reason for better performance.


\subsubsection{Finetuning tasks}

\medskip
\noindent\textbf{Question Answering}. We select the following tasks for finetuning: SQuAD v1.1\footnote{\url{https://raw.githubusercontent.com/rajpurkar/SQuAD-explorer/master/dataset/}} for English \citep{squad_paper}; FaQuAD\footnote{\url{https://raw.githubusercontent.com/liafacom/faquad/master/data/}} for Portuguese \citep{faquad2019}; GermanQuAD\footnote{\url{https://www.deepset.ai/datasets}} for German \citep{german-bert}; ViQuAD\footnote{\url{https://sites.google.com/uit.edu.vn/uit-nlp/datasets-projects}} for Vietnamese \citep{viquad_paper}.

\medskip
\noindent\textbf{Natural Language Inference}. We use: MNLI\footnote{\url{https://cims.nyu.edu/~sbowman/multinli/}} for English \citep{mnli_paper}; ASSIN2\footnote{\url{https://sites.google.com/view/assin2/}} for Portuguese \citep{assin2}; and XNLI\footnote{\url{https://github.com/facebookresearch/XNLI}} for both Vietnamese and German \citep{conneau2018xnli}.

\section{Results}

Table~\ref{table:results-qa} reports the results for the question answering experiments and Table~\ref{table:results-nli} for natural language inference. For every experiment, we provide a characterization of the pretraining (\textbf{source}) and finetune (\textbf{target}) languages we used. We also denote \textit{None} as the models of our control experiment, which are not pretrained.

We use a color scheme for better visualization. Results highlighted in \textcolor{red}{\textbf{red}} are considered the upper bound (models pretrained and finetuned on the same language); \textcolor{orange}{\textbf{orange}} is used for models finetuned on a different language from pretraining; \textcolor{blue}{\textbf{blue}} is used for lower bound experiments (control experiments).

\begin{table}[t]
\centering
    \begin{tabular}{ r | c | c | c | c }
    \hline
    \textbf{Pretraining} & \multicolumn{4}{c}{\textbf{Finetune and Evaluation}} \\
    & en & de & pt & vi \\
    \hline
    None & \textbf{\textcolor{blue}{21.50}} & \textbf{\textcolor{blue}{13.38}} & \textbf{\textcolor{blue}{27.23}} & \textbf{\textcolor{blue}{13.94}}\\
    en & \textbf{\textcolor{red}{88.32}} & \textbf{\textcolor{orange}{39.28}} & \textbf{\textcolor{orange}{45.22}} & \textbf{\textcolor{orange}{50.97}} \\
    de & \textbf{\textcolor{orange}{78.25}} & \textbf{\textcolor{red}{68.63}} & \textbf{\textcolor{orange}{43.85}} & \textbf{\textcolor{orange}{36.41}} \\
    pt & \textbf{\textcolor{orange}{79.31}} & \textbf{\textcolor{orange}{44.28}} & \textbf{\textcolor{red}{59.28}} & \textbf{\textcolor{orange}{55.99}} \\
    vi & \textbf{\textcolor{orange}{76.03}} & \textbf{\textcolor{orange}{29.56}} & \textbf{\textcolor{orange}{40.00}} & \textbf{\textcolor{red}{79.53}}\\

    \hline    
    \end{tabular}

    \caption{Results (F1 score) for question answering tasks. Results highlighted in red are considered the upper bound, while blue is used for lower bound. Only cells of the same column are comparable since results belong to the same task.}
    \label{table:results-qa}
\end{table}

\begin{table}[t]
\centering
    \begin{tabular}{ r | c | c | c | c }
    \hline
    \textbf{Pretraining} & \multicolumn{4}{c}{\textbf{Finetune and Evaluation}} \\
    & en & de & pt & vi \\
    \hline
    None & \textbf{\textcolor{blue}{61.54}} & \textbf{\textcolor{blue}{59.48}} & \textbf{\textcolor{blue}{70.22}} & \textbf{\textcolor{blue}{54.09}}\\
    en & \textbf{\textcolor{red}{83.85}} & \textbf{\textcolor{orange}{67.71}} & \textbf{\textcolor{orange}{82.56}} & \textbf{\textcolor{orange}{64.20}} \\
    de & \textbf{\textcolor{orange}{71.22}} & \textbf{\textcolor{red}{78.27}} & \textbf{\textcolor{orange}{77.45}} & \textbf{\textcolor{orange}{47.98}} \\
    pt & \textbf{\textcolor{orange}{75.56}} & \textbf{\textcolor{orange}{65.06}} & \textbf{\textcolor{red}{86.56}} & \textbf{\textcolor{orange}{63.15}} \\
    vi & \textbf{\textcolor{orange}{72.05}} & \textbf{\textcolor{orange}{66.63}} & \textbf{\textcolor{orange}{80.64}} & \textbf{\textcolor{red}{76.25}}\\

    \hline    
    \hline    
    \small{\textit{\# of classes}} & \small{3} & \small{3} & \small{2} & \small{3} \\
    \hline    
    \end{tabular}

    \caption{Results (accuracy) for Natural Language Inference in each language with different pretrainings. The last row highlights the number of classes for each task.}
    \label{table:results-nli}
\end{table}

\medskip
\noindent\textbf{Models leverage pretraining, even when finetuned on different languages.} For all selected languages, models can take advantage of its pretraining even when finetuned on a different language. If we compare, for instance, results on SQuAD (finetune and evaluation for English in Table~\ref{table:results-qa}), models pretrained in all languages outperform the control experiment by, at least, 50 points in F1 score. A very similar observation can be drawn from the results in other languages.

\medskip
\noindent\textbf{For a specific task, models that were pretrained on a different language achieve similar results}. An interesting observation is that results reported in a foreign language are quite similar. Taking SQuAD results, for instance, when using models pretrained on different languages, we see that F1 scores are around 78. The same can be observed in FaQuAD (finetune and evaluation for Portuguese in Table~\ref{table:results-qa}), for which results are close a F1 score of 43.


\medskip
\noindent\textbf{German-Vietnamese pairs have lower performance compared to other language pairs}. The intersection between German and Vietnamese contains the lowest performance among models finetuned on a foreign language. For question answering, we see that the Vietnamese model (BERT-VI) finetuned on German is 10 F1 points below the English pretrained model. The BERT-DE model also scores about 14 points lower in the ViQuAD task.
One explanation is that German is a more distant language to Vietnamese than English or Portuguese. However, since there is not a strong consensus in the linguistic community about the similarity of languages, it is not straightforward to test this hypothesis.

\section{Discussion}

Previous work has shown that using bi-lingual or multi-lingual pretrained models resulted in cross-lingual representations, enabling a model to learn a task in any of the languages used during pretraining. Our results show, however, that we can still achieve a good performance in a foreign language (i.e., a language not seen during pretraining).


Our experiments suggest that, during MLM pretraining, the model learns language-agnostic representations and that both language and task knowledge can be acquired during finetuning only, with smaller datasets compared to pretraining.

\subsection{Monolingual models performance in foreign languages}

The most interesting finding in this work is that a monolingual model can learn a different language during a finetune procedure. Differently from the literature, we use a model pretrained on a single language and targeted to a task in a different language (English models learning Portuguese tasks and vice versa).

We see that those models achieve good results compared to models pretrained and finetuned on the same language and largely outperform models without pretraining. This finding suggests that MLM pretraining is not only about learning a language model, but also concepts that can be applied to many languages. This had been explored in more recent work that leverages language models in different modalities \citep{lu2021pretrained}.

It is also noteworthy that those models could learn a different language with a finetune procedure even with small datasets (for instance, FaQuAD contains only 837 training examples; GermanQuAD has 11,518; and ViQuAD is comprised of 18,579). The English version of BERT could achieve a performance close to the upper bound for Portuguese on both NLI (82.56\% of accuracy) and QA (45.22 F1 score) tasks with small datasets. This, together with the low performance of control models, shows that the model leverage most of its pretraining for a task.

\subsection{Similar performance regardless of the pretraining language}

Another finding is that similar results were achieved by models pretrained on a different language than the task's. We can draw three different clusters from the results: the lower bound cluster, containing results for models without pretraining; the upper bound, with models pretrained on the same language as the task; and a third cluster, closer to the upper bound, with models pretrained on a different language.  

Since the models are comparable in pretraining data size and number of parameters, we hypothesize that this observation is related to the language-agnostic concepts learned during pretraining \textit{versus} language-specific concepts. The performance gap between the upper bound and third clusters may be explained by the language-specific properties not available to the latter group. 

This conclusion can also be drawn from the experiments performed by \citet{Artetxe2019}. The authors demonstrated that only swapping the lexicon enabled knowledge transfer between two languages. Since the embedding layer corresponds to a small portion of the total number of parameters of the model, it seems that language-specific properties represent a small portion of the whole model as well. Here, we go even further by showing that not even tokenizers and token embeddings need to be adapted to achieve cross-lingual transfer.

\section{Conclusion}

Recent work has shown that training a model with two or more languages and shared parameters is important, but also that representations from monolingual models pretrained with MLM can be aligned with simple mapping functions.

In this context, we perform experiments with monolingual models on foreign language tasks. Differently from previous work, we pretrain our models on a single language and evaluate their ability to learn representations for a task using a different language during finetuning, i.e., without using any cross-lingual transfer technique.

Our results suggest that models pretrained with MLM achieve language-agnostic representations that can be leveraged for target tasks. We argue that MLM pretraining contributes more to representations of abstract concepts than to a specific language itself.

The experiments conducted in this paper raise questions about the language-agnostic properties of language models and open directions for future work. Studying language-independent representations created from MLM pretraining may result in new ways to leverage this pretraining procedure for enabling models to perform in a wider range of tasks. More experiments on a wide range of languages and tasks are needed to strengthen our findings, which may lead to innovative language-agnostic methods for language models.

\section*{Acknowledgments}
This research was funded by a grant from Fundação de Amparo à Pesquisa do Estado de São Paulo (FAPESP) \#2020/09753-5.

\bibliographystyle{acl_natbib}
\bibliography{main}

\clearpage
\newpage





\end{document}